\documentclass[11pt,a4paper]{article}
\usepackage[hyperref]{acl2021}
\usepackage{times}
\usepackage{latexsym}

\usepackage{comment}
\usepackage{subcaption}
\usepackage{booktabs}
\usepackage{times}
\usepackage{latexsym}
\usepackage[T1]{fontenc}
\usepackage{tabularx}
\usepackage{multirow}
\usepackage{array}
\usepackage{amsmath}
\usepackage{graphicx}

\usepackage{microtype}

\definecolor{green}{RGB}{8, 180,30}
\definecolor{teal}{RGB}{0, 158, 115}
\definecolor{auburn}{RGB}{213, 94, 0}
\definecolor{orange}{RGB}{255, 127, 14}
\definecolor{blue}{RGB}{0,0,180}
\definecolor{purple}{RGB}{148, 103, 189}
\newcommand{\tb}[1]{\textcolor{blue}{#1}}
\newcommand{\tg}[1]{\textcolor{green}{#1}}

\newcommand{\taub}[1]{\textcolor{auburn}{#1}}

\definecolor{grey}{RGB}{100, 100, 100}

\usepackage{microtype}

\aclfinalcopy 

\makeatletter
\def\@fnsymbol#1{\ensuremath{\ifcase#1\or \dagger\or \ddagger\or
   \mathsection\or \mathparagraph\or \|\or **\or \dagger\dagger
   \or \ddagger\ddagger \else\@ctrerr\fi}}
    \makeatother
   
\title{InFillmore: Frame-Guided Language Generation \\ with Bidirectional Context}
\author{
Jiefu Ou\thanks{~~Corresponding authors.}~\thanks{ \ \ Work done during an internship at the Center for Language and Speech Processing, JHU.}\quad  
Nathaniel Weir\footnotemark[1]~\quad  
Anton Belyy\footnotemark[1]~\quad  
Felix Yu\quad  
Benjamin Van Durme\\
Johns Hopkins University \\
  {\tt  jouaa@connect.ust.hk}\\
  {\tt \{nweir,abel,fyu17,vandurme\}@jhu.edu}
  }

\date{}

\begin{document}
\maketitle
\begin{abstract}
We propose a structured extension to bidirectional-context conditional language generation, or ``infilling,'' inspired by Frame Semantic theory~\cite{fillmore1976frame}. 
Guidance is provided through two approaches: (1) model fine-tuning, conditioning directly on observed symbolic frames, and (2) a novel extension to disjunctive lexically constrained decoding that leverages frame semantic lexical units. 
Automatic and human evaluations confirm that frame-guided generation allows for explicit  manipulation of intended infill semantics, with minimal loss in distinguishability from human-generated text. Our methods flexibly apply to a variety of use scenarios, and we provide an interactive web demo.\footnote{Codebase and demo available from  \url{https://nlp.jhu.edu/demos/infillmore}.}

\end{abstract}
\graphicspath{{./figures/}}

\section{Introduction}

A popular strategy for automatic story generation is to proceed in a coarse-to-fine manner: first by proposing a \textit{story plan}, and then realizing it into natural language form using large pretrained neural language models~\cite{fan2018hierarchical,goldfarb-tarrant-etal-2019-plan}. In this work, we study the use of FrameNet frames~\cite{baker1998berkeley} as representational units for such plan guidance.  

In Frame Semantics~\cite{fillmore1976frame,Fillmore2009AFA}, words evoke structural situation types (frames)
that describe the common schematic relationships between lexical items.
We hypothesize that these structured types can be used to effectively induce the semantic content of text generated by increasingly powerful pretrained language models,
yielding a flexible, controllable and domain-general model for surface realization of story plans with a variety of dimensions for user guidance.

\begin{figure}[t!]
    \centering
    \includegraphics[width=.39\textwidth]{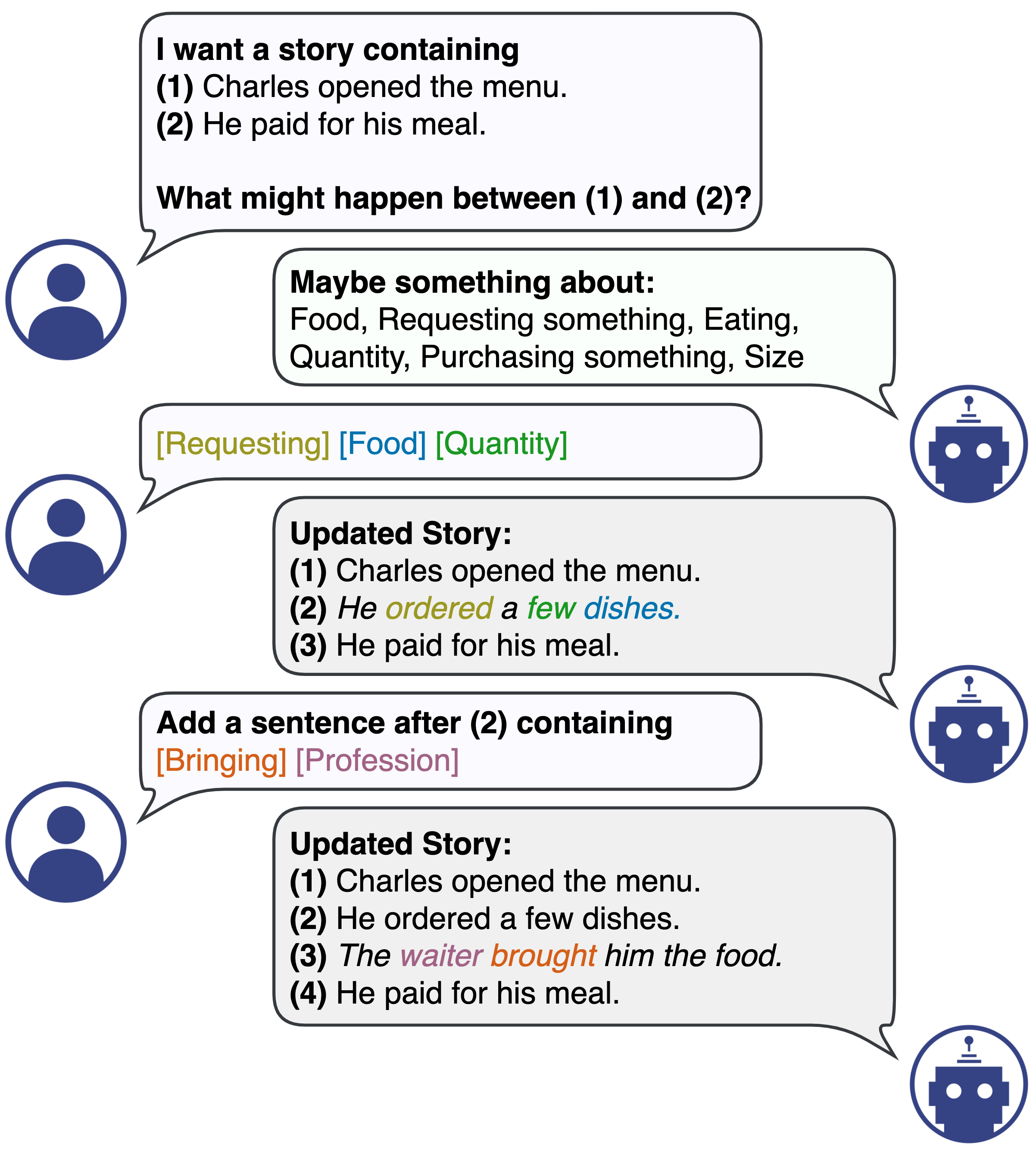}
    \caption{The proposed generation model, applied to the interactive story generation task. Similar to the existing infilling models, a user can insert or rewrite text spans at any position in a story. With the proposed extension, generation can be guided via explicit frame semantic constraints, either provided manually or suggested by the model based on surrounding context.}
    \label{fig:overview}
\end{figure}

Based on this supposition, we fine-tune a recent infilling model~\cite{donahue2020enabling} with a frame-guided denoising objective.
We contrast this approach with a novel method for frame-guided generation that modifies only the decoding step of a standard language model through lexical manipulation. The idea originates from the annotation scheme of FrameNet, where each semantic frame is annotated with a set of evocative lexical units (LUs).
We posit that it is possible to guide the model's generation with frames \textit{without} modifying its training procedure by instead lexically constraining its generation output to contain frame-associated LUs. 
Therefore, we develop an extension 
to lexically-constrained decoding that leverages LUs as ordered disjunctive constraint sets. Given a possibly multi-frame sequence 
and a generative model, our method enforces the generation of one of the associated LUs for each frame in the sequence. This decoding method is implemented as a plug-and-play module that can be imposed on top of any standard generative language model.\footnote{Fairseq-based implementation and data to be released.}

We evaluate through a sentence-infilling task based on  ROCStories~\cite{mostafazadeh2016corpus}, 
assessing performance on two dimensions: 1) the \emph{quality} of generation,  as measured through perplexity   
and human evaluation; and 2) the \emph{fidelity}, which scores whether generated text evokes the frames used as guidance.
We demonstrate that our methods utilize guidance to generate frame-evoking surface realizations without meaningfully detracting from the contextual narrative coherence. 
We also demonstrate the practical applicability of frame-guided generation in a variety of example use cases. 

\section{Related work}
\paragraph{Controlled Generation}
Existing work employs a variety of pretraining strategies to guide and/or diversify text generation.  \newcite{keskar2019ctrl} train large-scale language models on text prepended with \textit{control codes}, allowing for guided content and style. PPLM~\cite{dathathri2019plug} makes use of lightweight \textit{attribute classifiers} that guide generation without requiring language model retraining.
For diverse generation of sentences in a more general scenario, \newcite{weir-etal-2020-cod3s} train models to condition on \textit{semantic bit codes} obtained from hashing sentence embeddings.

\paragraph{Constrained Generation}
Separate lines of work employ \textit{lexical constraints} to achieve the same goal of guided and diverse generation. As such, 
lexically constrained beam search methods such as Grid Beam Search \cite{hokamp2017lexically} and Dynamic Beam Allocation \cite{post2018fast, Hu2019ImprovedLC} were proposed as the
decoding methods for causal generation with disjunctive positive constraints \cite{li2020guided}, paraphrasing \cite{Hu2019ParaBankMB, Culkin2020IterativePA}, machine translation \cite{Zhang2019NeuralMT}, and abstractive summarization \cite{Mao2020ConstrainedAS}.
\newcite{lu2020neurologic} generalize beam-search based methods with an algorithm that supports lexical constraints in the conjunctive normal form. 

Parallel are the approaches that handle lexical constraints in an editing manner: starting with a sequence of keyword constraints and fleshing out a sentence via editing operations such as insertion or deletion
~\cite{Miao2019CGMHCS,Liu2019BFGANBA,  Sha2020GradientguidedUL,Susanto2020LexicallyCN,Zhang2020POINTERCP}.
Finally, it is possible to satisfy lexical constraints in a soft manner as external memories \cite{Li2020NeuralMT, Li2019LearningEL} or constructing constraint-aware training data \cite{Chen2020LexicalConstraintAwareNM}.

\paragraph{Story Generation} Inspired by the traditional pipeline of \newcite{reiter2000building}, 
recent work tackles generation of stories in a coarse-to-fine manner~\cite{fan2018hierarchical}: based on a premise, a structured \textit{outline} is generated first, and then an outline-condition model generates the full story. To represent the story outline, existing approaches typically either model it as a latent variable, or use symbolic representations such as key phrases \cite{xu2018skeleton,yao2019plan,goldfarb-tarrant-etal-2019-plan,gupta2019writerforcing,rashkin2020plotmachines}, short summaries \cite{jain2017story,chen2019learning}, verb-argument tuples \cite{martin2017event}, or PropBank predicates and arguments  \cite{fan2019strategies,goldfarb2020content}. Our work can be viewed as an extension of this direction, where a \textit{Content Planner} model generates an outline as a sequence of FrameNet frames, and our methods generate a surface form story.


\section{Data}
\paragraph{FrameNet}
FrameNet is a 
lexical database of English
based on Fillmore's theory of Frame Semantics. It defines more than 1200 frames spanning various semantic domains, where each frame schematically describes a type of event, relation, or entity. A frame is defined with a set of corresponding \textit{Frame Elements} (FEs): the participants in the frame with relational roles, and a set of \textit{Lexical Units} (LUs): words that evoke the frame in text.

For example, the \textbf{Apply\_heat} frame that describes the concept of cooking consists of core FEs \textit{Food}, \textit{Cook}, \textit{Container}, \textit{Heating\_instrument}, and \textit{Temperature\_setting}, and has evocative LUs that include \textit{fry, bake, boil}, and \textit{broil}.
Frame annotations provide a partial (albeit rich) picture of sentence meaning, i.e. information not governed by the syntax/semantics interface. We find that they serve as an effective, theory-grounded formalism for discrete semantic guidance of generation. 

Conceptually, our choice to use FrameNet as guiding semantics builds upon trends in generative modeling of discourse~\cite{ferraro2016aub} that treat text documents as mixtures of hierarchical latent variables in accordance with classical theories of frame semantics (e.g. \newcite{minsky1974framework,fillmore1976frame}). As described by \newcite{ferraro2016aub},  FrameNet frame information can be used to learn a hierarchical latent representation of sentence-level
semantics 
that produces discourse models that better fit to natural text data.
Our work then asks whether this information can be used to harness the increasingly powerful ability of recent neural language models for the purposes of controlled story generation. 
\begin{figure}[t!]
\small
    \centering
    \setlength\tabcolsep{3pt}
    \begin{tabular}{rl}
    \toprule
    \textbf{Story} & 
    \multicolumn{1}{l}{Charles went shopping. He bought fruit. } \\ 
    & Then he left.\\ \midrule 
     \textbf{ILM} 
         & Charles went shopping. \textcolor{blue}{[blank]} Then he left. \\ 
      &  \textit{\tb{[sep]} \tg{He bought fruit.}} \\ 
     \textbf{S-FFL}    
     &   \tb{[sep] [Food]} \tg{He bought fruit.}  \\
     \textbf{A-FFL} 
     &   \tb{[sep] [Commerce\_buy] [Food]} \tg{He bought fruit.}  \\
     \bottomrule
    \end{tabular}
    \caption{Training examples for frame-guided ILM models. Examples are fed from left to right, with the \textit{italicized} portion of the ILM example replaced by the frame-injected sequences for FFL examples. }
    \label{fig:ffl-training}
\end{figure}
\paragraph{ROCStories}
\citet{mostafazadeh2016corpus} introduce the ROCStories corpus, which comprises over 98K 5-sentence simple stories that can serve as a resource for commonsense narrative schema learning and story generation \cite{ippolito-etal-2020-toward}. We use this dataset
to evaluate the performance of our methods (described in \autoref{subsec:setup}).

\section{Approach}
\subsection{Model Architecture}
The Infilling by Language Modelling \citep[ILM,][]{donahue2020enabling} framework fine-tunes pretrained unidirectional language model such as GPT-2 \cite{radford2019language}
to generate target infill spans with bidirectional contexts.
This allows the ILM model to flexibly generate text at any position in a document, as shown in \autoref{fig:overview}.
In this work, we introduce FrameNet frame guidance into the ILM pipeline. 
We propose and compare methods based on 1) fine-tuning on frame-annotated data (\ref{subsec:FFL}), and 2) imposing lexically-constrained beam search during decoding (\ref{subsec:LCD}) with the original ILM.

\subsection{Fine-Tuned ``Framefilling'' (FFL)}\label{subsec:FFL}
The ILM task definition comprises a context passage ${x}$ containing \textbf{[blank]} tokens at points where the new spans must be generated.\footnote{Our work focuses primarily on infilling single sentence spans, leaving arbitrary length spans, e.g. $n$-grams or full paragraphs, to the future work.} 
The passage $x$ is concatenated with a \textbf{[sep]} token and golden span infills (each separated by another \textbf{[sep]}) to form a fine-tuning instance for an off-the-shelf unidirectional language model such as GPT-2. We build on this setup by adding one or more frame ID tokens $F_1,F_2,\dots$ (e.g. \textbf{[Food]}) as prefixes of each golden infill span, as shown in \autoref{fig:ffl-training}. A model fine-tuned on this modified formulation, which we call a ``framefilling'' model (FFL for short), therefore conditions each infill on the bidirectional context as well as one or more control codes that guide the infill's semantic content. If an example contains multiple infills, subsequent infills are conditioned on the frames and text of previous infills.

We experiment with multiple variants of the FFL model, varying primarily in the level of frame guidance. 
We train a variant on infilling examples that contain a single frame ID (S-FFL), another on examples with a set of one or multiple frames (M-FFL; number of frames sampled from a geometric distribution with $p=.4$), and a final variant conditioned on all frames (covered by FrameNet v1.7) triggered by the infill (A-FFL). In all cases, the frame ID tokens are predicted by a state-of-the-art neural FrameNet parser~\cite{xia-etal-2021-lome}.\footnote{We choose to evaluate these three variants in order to compare the coherence of a model trained with only low frame guidance (S-FFL), a model trained with only high (A-FFL), and a model (M-FFL) trained on a distribution of examples that comprises a superset of the first two. }

\begin{figure}
    \centering
    \includegraphics[width=1.0\columnwidth]{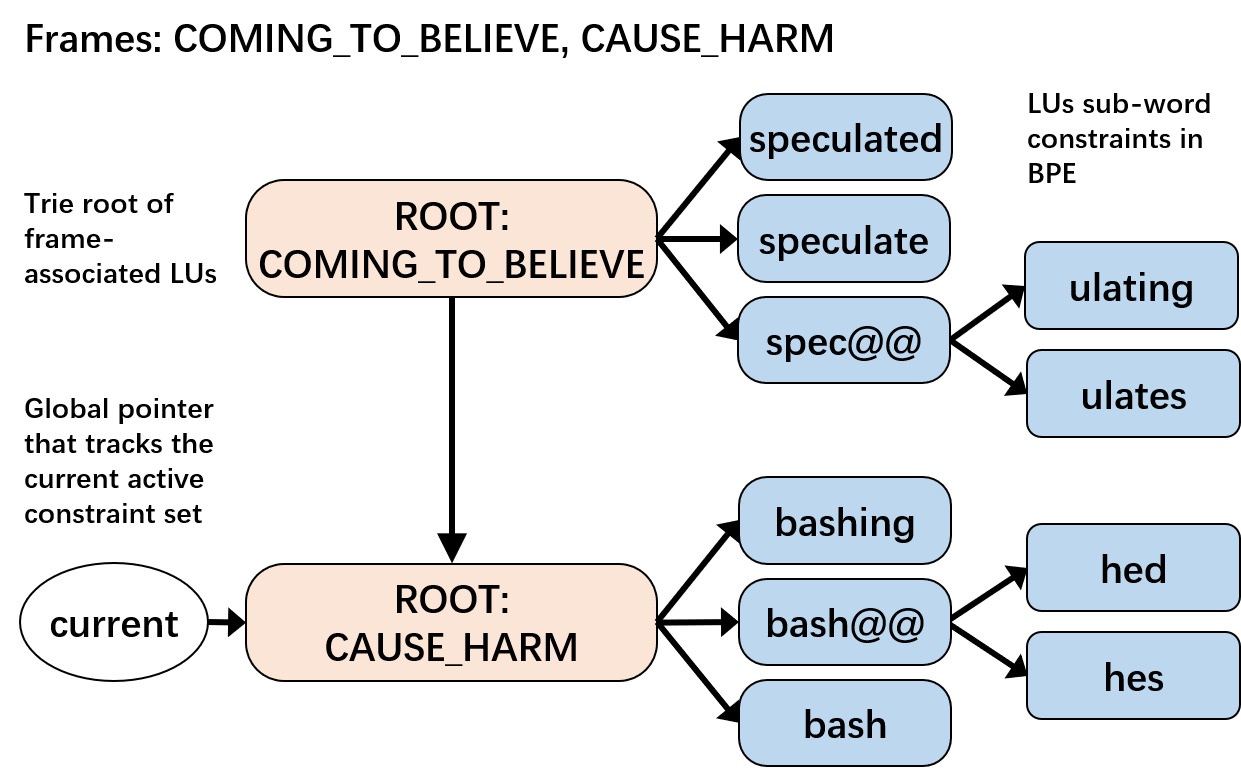}
    \caption{Example of LCD constraint representation in a list of 2 tries, corresponding to the given frames \textbf{Coming\_to\_believe} and \textbf{Cause\_harm}. Other LUs are omitted for simplicity.}
    \label{fig:lcd_tries}
\end{figure}
\subsection{Lexically Constrained Decoding (LCD)}\label{subsec:LCD}

Given a sequence of frame ID tokens $F_1, F_2, ..., F_n$, we build a corresponding sequence of disjunctive lexical constraint sets $C_1, C_2, ..., C_n$, where $C_i$ consists of all LUs of $F_i$ with their morphological variants. During decoding, our method forces the output to contain $c_1, c_2, ... c_n$, where $c_i \in C_i$.

\paragraph{Decoding with 
Ordered/Unordered 
Disjunctive Constraint Sets} 

We develop a disjunctive lexically constrained decoding method ($\textbf{LCD}$) that extends implementations in \newcite{post2018fast, Hu2019ImprovedLC} and \newcite{li2020guided}. We also use Dynamic Beam Allocation (DBA) \cite{post2018fast, Hu2019ImprovedLC} for beam assignment and next token selection, but we track our constraints differently.
As shown in \autoref{fig:lcd_tries}, LCD represents a sequence of disjunctive constraint sets as a list of tries, one per frame, each covering a set of disjunctive lexical units (with morphological variants) based on the Byte Pair Encoding \citep[BPE,][]{sennrich2016neuralmt} adopted by GPT-2. 

Based on this representation, we develop two versions of LCD: LCD-ordered and -unordered, the former of which requires that the constraint sets be completed in the order that the corresponding frame ID tokens are specified. By providing these two versions, we offer the user the flexibility to either enforce the frame-evoking narration being triggered in their desire order, or leave it to be determined by the generative model and decoder.

To track the generation progress through constraint sets, we use a global pointer to the currently active disjunctive set. Whenever the active set $C_i$ is completed, the pointer is set to null. If unsatisfied sets remain, the next possible set(s) to be completed is $C_{i+1}$ for LCD-ordered  and $\{C_j: j \neq{i} \in \{1, 2, ..., n\} \ | \  C_j \ \text{is not completed}\}$ for LCD-unordered. At the beginning of generation when no set is active, the next possible set(s) is $C_{0}$ for LCD-ordered and all sets for LCD-unordered. During the generation, when the pointer is null and a constraint token that starts any of the next possible set(s) is picked by DBA, the global pointer is set to the corresponding disjunctive set. 
Apart from the global pointer, the bookkeeping and unwinding mechanism within each trie is similar to the implementations in \citep{Hu2019ImprovedLC} and \citep{li2020guided}, except that a trie is marked as finished and the global pointer is updated once any path in the trie is completed.

We implement LCD as an extension of the token generation constraint implementation in the \texttt{fairseq} library.
Our LCD works very similarly to the disjunctive positive constraints decoding in \cite{li2020guided}, where the disjunctive sets are maintained in a single trie rather than our ``list of tries'' approach. However, we support explicit ordering of constraint sets, and we don't prune a sub-trie when the corresponding constraint set is finished.

\section{Experiments}
\label{subsec:setup}
\vspace{-.1cm}
We test the effectiveness of our models on a frame-guided sentence infilling task derived from ROCStories. We use a state-of-the-art neural FrameNet parser~\cite{xia-etal-2021-lome}
to obtain the set of frames evoked by each sentence in the dataset.
We then present models with a five-sentence ROC story with one masked out. The model must infill the missing sentence given one or many frame ID tokens parsed from the masked-out sentence. For evaluations requiring generated outputs (all but perplexity), we use beam search with beam size 20. We find that beam search achieves higher frame fidelity and coherence than the random sampling approach used by \newcite{donahue2020enabling}.

We train our models (all GPT-2 `base') using the provided train split of ROCStories. For S/M/A-FFL, each example contains one/multiple/all frame ID tokens sampled randomly from the parser output. 
To test LCD, we re-train the original ILM 
using the identical ROCStories training data to our FFL models but  without frame tokens (training details described in \ref{app:ILM}). Unlike \newcite{donahue2020enabling}, we do not include story titles.
We also use this ILM as a baseline with no guidance.

To investigate whether enforcing generated frame order impacts model performance, we evaluate both LCD-ordered  and -unordered; we also evaluate FFL-ordered models fine-tuned to generate frames in the order in which they are provided.  

\subsection{Automatic Evaluation}
\label{sec:auto-eval}
We evaluate our frame-guided generation methods by measuring the rate at which they produce sentences that trigger the desired frame(s) and by measuring the perplexity score of the framefilling-trained language model on test examples. 
\paragraph{Frame Fidelity}
We automatically evaluate whether a produced sequence triggers a given set of frames by running it through the same neural frame parser used to determine the desired frame from a gold human-generated sentence. \autoref{tab:fidelity} shows
the rates at which methods correctly produce sentences that contain every specified frame.\footnote{Methods that condition on fewer frame IDs are evaluated using subsets of those for multi-frame models; e.g. if the M-FFL model must generate a set \textbf{\{[Food] [Size]\}}, the S-FFL must predict one of \textbf{[Food]} or \textbf{[Size]}. } For each model, we evaluate the top-1 decoded sequence.
\begin{table}[t!]
    \centering
    \small
    \setlength\tabcolsep{3pt}
    \begin{tabular}{lccccc}
    \toprule
    \multirow{1}*{\textbf{Fidelity} $\uparrow$} & \multicolumn{3}{c}{\textbf{Recall}} & \multicolumn{2}{c}{\textbf{Perfect Recall}} \\  \cmidrule(lr){2-4}\cmidrule(lr){5-6}
     \multicolumn{1}{r}{\textbf{\# Frames}} &   \textbf{Single} & \textbf{Multi} & \textbf{All} & \textbf{Multi} & \textbf{All}   \\ \midrule 
    ILM (no guidance)     & .169          & .166          & .165 & .091 &  .026 \\
    ILM + LCD             & \textbf{.584} & .595          & .610 & .418 & .232 \\
    ILM + LCD-ord     & --            & \textbf{.598} & .626 & \textbf{.427} & .255  \\
    FFL                   &  .518         &     .559      &  .640 & .381 & .259 \\
    FFL (rand sample) &  .461         & .511          & .601 & .338 &  .224  \\ 
    FFL-ord         & --            & .585          &  \textbf{.669} & .415 &  \textbf{.298}  \\
     \bottomrule
    \end{tabular}
    \caption{Frame fidelity, computed as frame recall according to the neural frame parser (left). The per-example rate at which models perfectly predict frame sets is also given (right). Higher is better.
    }
    \label{tab:fidelity}
\end{table}

\paragraph{Perplexity}

The typical automatic evaluation metric for a language model is test data perplexity (PPL). Since LCD requires no training modification to the ILM model, we only compute the PPL for S/M/A-FFL and ILM on a test set of stories in which one of five sentences has been masked out.\footnote{See \autoref{app:ppl} for model perplexity trained and evaluated with all five sentences having been masked.} 
Following \citet{donahue2020enabling}, we evaluate models' PPL specifically on infill tokens and also compute PPL including the surrounding special tokens (separators and frame IDs). 
Because sequences for FFL models include one or more frame ID tokens, the token length for a given story example is different for ILM and each FFL variant; PPL therefore cannot be directly compared. 
To construct a scenario in which the ILM and FFL model perplexities \textit{are} directly comparable, we train variants of both models for which every infill sequence is prepended with 5 special tokens, thus regularizing token length for every evaluated model.

\subsection{Human Evaluation}
In addition to automatic evaluation, we collect human judgements to assess models' ability to maintain coherent and plausible generation.
We conduct two human evaluations that ask annotators to tell apart model- and human-generated sentences (Indistinguishability task) and rank model-generated sentences relative to one another (Relative Plausibility task). Details of our collection protocols and example interfaces are provided in \autoref{app:human-eval}.

\paragraph{Indistinguishability}
Following \citeauthor{donahue2020enabling}, we present annotators with 5-sentence stories in which one sentence has been replaced by the output of an infilling model. Annotators must identify which sentence is model-generated.

For each model, we calculate the confusion rate $r = \frac{N_{\textit{confused}}}{N_{\textit{all}}}$, where $N_{\textit{confused}}$ is the number of stories for which a human annotator fails to identify the machine-generated content, and $N_\textit{all}$ is the total number of stories. Results are shown in \autoref{tab:indistinguish}. Higher confusion rate is posited to mean more natural text infilling. Optimal performance is $80\%$, meaning the annotator is performing at chance.

\paragraph{Relative Plausibility}
We present human annotators with a 5-sentence story where one sentence is missing, and 10 candidate replacement sentences (the gold plus the infills of 9 different models). Annotators are tasked with ranking the candidate sentences (via drag-and-drop) based on how plausible they are relative to each other. 
Upon aggregating judgements, each model's score is calculated as the average relative rank of its output sentences that are assigned by annotators, as shown in \autoref{tab:rel-plau-rank}.

\begin{table}[t!]
    \small
    \centering
    \setlength\tabcolsep{3pt}
    \begin{tabular}{lrrrrr} \toprule
    \multirow{2}*{\textbf{Perplexity} $\downarrow$ } & \multirow{2}*{ILM} & \multirow{2}*{S-FFL} & \multirow{2}*{M-FFL} & \multirow{2}*{A-FFL} & A-FFL  \\ &&&&& (ord) \\ \midrule
        Infill Text  & 12.85  &  11.7 & 9.84 & 6.19 & \textbf{5.05} \\
        \quad + Sp Toks & 7.24 & 8.32 & 9.5 & 8.95 & \textbf{7.04}             \\
        \quad w/ 5 Fr Slots & \textbf{4.06} & 5.12 & 6.34 & 7.32 & 6.03 \\
        \bottomrule
    \end{tabular}
    \caption{Model perplexity over infill text tokens and infill text tokens + special tokens (\textless start to infill\textgreater, \textless end of infill\textgreater, \textless infill mask\textgreater). Lower is better.}
    \label{tab:ppl}
\end{table}
\section{Analysis}

\paragraph{Fidelity}
From the results in \autoref{tab:fidelity}, we find that ILM+LCD, FFL and FFL-ordered all perform similarly while substantially outperforming the baseline unguided ILM. This shows that our methods
effectively produce text evoking the desired frame semantic content.
Both methods benefit from the inclusion of gold frame order, more so for FFL.

There is a considerable gap between the performance of our models and perfect performance (1.0). This is because FFL operates only with soft ``control code'' constraints, and although LCD is strictly required to generate trigger LUs for every frame, it does not produce sentences that always successfully evoke the frame. While some of this gap might be the result of imperfections of the parser,
we find word sense ambiguity to be a contributing problem. Many LUs, such as \textit{work.v}, \textit{see.v}, or \textit{call.v} have multiple senses each associated with a different frame. Since neither LCD nor FFL imposes hard constraints on word sense, it is entirely possible for an unintended sense to be generated.

\begin{table}[t]
    \centering
    \setlength\tabcolsep{4pt}
    \begin{tabular}{llll}
    \toprule
    \multirow{2}*{\textbf{Confusion rate (\%) $\uparrow$}} & \multicolumn{3}{c}{\textbf{\# Frames}} \\  \cmidrule(lr){2-4}
    & \textbf{Single} & \textbf{Multi} & \textbf{All}  \\ \midrule
     ILM (no guidance) & \textbf{41} & \textbf{41} & \textbf{41} \\ 
     ILM+LCD & 35 & 31$^*$ & 20$^*$ \\ 
     FFL & 33$^*$ & 39 & 38 \\ 
     FFL-ordered & 33$^*$ & 38 & 37 \\ 
      \bottomrule
    \end{tabular}
    \caption{Confusion rate computed as the 
    percentage of stories for which the annotator picks a wrong sentence as machine-generated. Higher is better. $^*$ denotes significant difference from the baseline (ILM) result, according to the two-sided McNemar test with $p < 0.05$.}
    \label{tab:indistinguish}
\end{table}

\begin{figure}[t!]
\small
    \centering
    \setlength\tabcolsep{3pt}
    \begin{tabular}{l}
    \toprule
    \textbf{Story}  \\
    \multicolumn{1}{l}{ $\cdots$ I danced terribly and broke a friend's toe. \tb{[blank]}} \\
    \textbf{Frame:} \tb{[Request]}   \\
    \midrule
    \textbf{Gold} \hspace{62pt} \textcolor{blue}{[Request]}\\
    \multicolumn{1}{l}{The next weekend, I was \textbf{\tb{asked}} to please stay home.} \\
    \midrule
    \textbf{ILM+LCD} \textcolor{red}{[Contacting]} \\
    \multicolumn{1}{l}{I went home to \textbf{\textcolor{red}{call}} my friend and tell her I broke her toe.}\\
    \bottomrule
    \end{tabular}
    \caption{From the lexical constraints on the LUs of the frame \textbf{Request} (as \textit{asked.v} in the gold infill), the decoder selects \textit{call.v}, 
    but in the generated context \textit{call} becomes a surfaces realization of frame \textbf{Contacting}.}
    \label{fig:lcd_fidelity}
\end{figure}

As illustrated in \autoref{fig:lcd_fidelity}, LCD forces picking the LU \textit{call.v} for the target frame \textbf{Request}, but given the subsequent output \textit{call my friend to tell her I was hurt}, the \textit{call.v} unit takes on a sense that triggers the incorrect frame \textbf{Contacting}.

\paragraph{Perplexity}
\autoref{tab:ppl} shows that the perplexity over purely the infill tokens is inversely proportional to the amount of frame guidance provided to the language model.
However, we find that under the directly comparable 5 slot scenario, PPL computed over the infill tokens plus all surrounding special/frame tokens is \textit{worse} for models with more frame tokens. As this work is predominantly concerned with the quality of generation given gold frame IDs, this is less of a concern; that the perplexity of infill tokens decreases considerably with the introduction of frame guidance shows that neural language models can be explicitly guided towards specific semantic spaces in accordance with the conceptual semantic structures underpinning human understanding of language.
\paragraph{Generation Quality}
\autoref{tab:rel-plau-rank} shows that in terms of human-judged relative plausibility, FFL outperforms all other models (including the unconstrained ILM) when conditioning on all frames, and underperforms ILM with only a small margin with multi-frame guidance. 
\autoref{tab:indistinguish} shows that ILM outperforms FFL models and LCD on the Indistinguishability task in all cases, but with only a small margin in multi/all-frame cases comparing with FFLs.  This is unsurprising, as ILM is optimized to replicate human-produced text under no constraints via semantic guidance. 
We observe as in the fidelity evaluation that LCD slightly outperforms FFL under single frame constraints in both human evaluations.
\begin{table}[t]
    \centering
    \setlength\tabcolsep{4pt}
    \begin{tabular}{llll}
    \toprule
    \multirow{2}*{\textbf{Average rank (1..10) $\downarrow$}} & \multicolumn{3}{c}{\textbf{\# Frames}} \\  \cmidrule(lr){2-4}
    &   \textbf{Single} & \textbf{Multi} & \textbf{All}  \\ \midrule
     ILM (no guidance) & \textbf{5.48} & \textbf{5.48} & 5.48 \\ 
     ILM+LCD & $5.85^*$ & 6.38$^*$ & 7.50$^*$ \\ 
     FFL & $5.88^*$ & 5.57 & 5.11 \\ 
     FFL-ordered & $5.88^*$ & 5.53 & \textbf{5.02$^*$} \\ 
     \bottomrule
    \end{tabular}
    \caption{Average relative plausibility rank by human annotators. Lower is better. $^*$ denotes significant difference from the baseline (ILM) result, according to the two-sided Wilcoxon signed-rank test with $p < 0.05$.}
    \label{tab:rel-plau-rank}
\end{table}
\begin{figure*}[t!]
    \centering
    \footnotesize
    \setlength\tabcolsep{4pt}
    \begin{tabular}{p{6cm}|p{9cm}}
    \toprule
    \begin{tabular}[t]{l}
     \textbf{Story}      \\
     Ari spends \$20 a day on pickles. \\
     He decides to make his own to save money. \\
     He puts the pickles in brine. \\
      \tb{[blank]} \\
      Ari opens the jar to find perfect pickles. \\
      \midrule 
      \textbf{Gold} \\
      Ari waits 2 weeks for his pickles to get sour. \\ \midrule
      \textbf{ILM Baseline} \\ 
      He puts the pickles in a jar. 
    \end{tabular}     &  
    \begin{tabular}[t]{ll}
    \multicolumn{1}{c}{\textbf{FFL}} & \multicolumn{1}{c}{\textbf{ILM+LCD}}  \\ \cmidrule(lr){1-1} \cmidrule(lr){2-2}
     \multicolumn{2}{l}{\textbf{Single Frame:} \tb{[Transition\_to\_State]}} \\ \cmidrule(lr){1-2}
     He ends up with a jar full of pickles. & He \tb{gets} the pickles and \\
     & puts them in jars.\\
     \midrule 
     \multicolumn{2}{l}{\textbf{Multiple Frames:} \tb{[Cardinal\_Numbers] [Transition\_to\_State]}} \\ \cmidrule(lr){1-2}
     He ends up with 5 jars of pickles. & He puts \tb{one} in the jar and opens  \\
     & it to \tb{get} a drink. \\ \midrule 
     \multicolumn{2}{l}{\textbf{All Frames:} \tb{[Cardinal\_Numbers] [Measure\_duration]}} \\
     \multicolumn{2}{l}{\hspace{1.65cm} \tb{[Transition\_to\_State][Chemical-sense\_description]}} \\ \cmidrule(lr){1-2}
     He waits for a week for the & He \tb{waits} for the pickles \\
     pickles to get sour. &  to thaw out of the jar to thaw\\
     & \tb{one} \tb{day} he \tb{gets} the pickles and \\
     & eats them \tb{delicious}.
    \end{tabular}
    \\ \bottomrule
    \end{tabular}
    
    \caption{Example infills by FFL, LCD and ILM baseline under single, multiple, and all frame guidance. Under single frame guidance, all decoding methods perform interchangeably. As the number of frames increases, FFL approaches a surface realization of frame-specified semantic content that resembles that of the gold infill. The unguided baseline ILM generates something relatively incoherent. 
    Under ``all frame'' guidance, LCD fails to satisfy all constraints in one sentence and generates an additional sentence that corrupts quality. 
    }
    \label{fig:ffl_lcd_trend}
\end{figure*}
\begin{figure*}[t!]
    \centering
    \footnotesize
    \setlength\tabcolsep{3pt}
    \begin{tabular}[t]{p{.31\textwidth} |  p{.31\textwidth} p{.31\textwidth}}
        \toprule
        \addtolength{\tabcolsep}{-2pt}
        \begin{tabular}[t]{p{.30\textwidth}}
            \textbf{A. Iterative Refinement} \\\cmidrule(lr){1-1}
            \textbf{User:} (I) Alice went to the grocery store. (II) \tb{[Commerce\_buy]} \\
            \textbf{System:} 
            (IIA) She bought all the ingredients for a cake. 
            (IIB) She bought a new pair of shoes. 
            (IIC) She bought a lot of fruits and veggies. \\
            \textbf{U:} \taub{Choose (IIA) as (II) and infer content after (II)} \\
            \textbf{S:} \tb{[Food]}, \tb{[Cooking]}, \tb{[Ingredients]}, \\\tb{[Desirability]}, \tb{[Time\_Collocation]} \\
            \textbf{U:} \taub{Choose} \tb{[Desirability] [Cooking]} \\
            \textbf{S:} 
            {(I) Alice went to the grocery store. 
            (II) She bought all the ingredients for a cake. }
            \textit{(III) She made the best cake she ever had.} \\
            \textbf{U:} \taub{Infer content to replace (I)} \\
            \textbf{S:} \tb{[Food]}, \tb{[Deciding]}, \tb{[Social\_Event]}, \\ \tb{[Building]}, \tb{[Quantity]} \\
            \textbf{U:} \taub{choose} \tb{[Social\_Event]} \\ 
            \textbf{S:} \textit{(I) Mary wanted to make a cake for her birthday. } {(II) She bought all the ingredients for a cake. 
            (III) She made the best cake she ever had.} \\
            \textbf{U:} \taub{Insert sentence about} \tb{[Motion]} \taub{at (II) and sentence about} \tb{[Temporal\_Collocation]} \taub{at (IV)} \\
            \textbf{S:} {(I) Mary wanted to make a cake for her birthday.} \textit{(II) She went to the store.} {(III) She bought all the ingredients for a cake.}  \textit{(IV) That afternoon, she baked the cake in the oven.} {(V) She made the best cake she ever had. }
        \end{tabular} 
        & 
        \multicolumn{2}{c}{
        \begin{tabular}[t]{p{.6\textwidth}}
            \addtolength{\tabcolsep}{-3pt}
            \begin{tabular}[t]{p{.3\textwidth}p{.3\textwidth}}
                \begin{tabular}[t]{p{.29\textwidth}}
                    \textbf{B. Generation from Skeleton} \\ \cmidrule(lr){1-1} 
                    \textbf{User:} (I) Bob went to Paris. (II) \tb{[Buildings]} (III) \tb{[Experience]} (IV) \tb{[Motion]} (V) \tb{[Calendric\_Unit]}\\
                    \textbf{System:} {(I) Bob went to Paris.}
                    \textit{(II) He went to see the Louvre museum.
                            (III) He saw lots of paintings and sculptures.
                            (IV) He went to see the Eiffel Tower.
                            (V) It was the best day of Bob's life.} \\
                \end{tabular} & 
                \addtolength{\tabcolsep}{6pt}
                \begin{tabular}[t]{p{.29\textwidth}} 
                    \textbf{U:} (I) Emma loved writing and wanted to be a writer. (II) \tb{[Deciding]} (III) \tb{[Practice]} (IV) \tb{[Publishing][Text]} (V) \tb{[Fame]}\\
                    \textbf{S:} {(I) Emma loved writing and wanted to be a writer.}
                    \textit{(II) She decided to enter a writing contest. 
                            (III) Emma's practice paid off and she won first prize. 
                            (IV) Emma was so proud of herself that she published a book. 
                            (V) Now Emma is a famous author.} \\
                \end{tabular} 
            \end{tabular} \\
            \midrule 
            \textbf{C. Diverse Candidate Generation} \\ \cmidrule(lr){1-1}
                    \textbf{User:} \taub{Infer candidates after} (I) Charles slipped on a banana peel.   \\
                    \textbf{System:} (IIA \tb{[Body\_Parts]})   He skinned his knee badly. (IIB \tb{[Motion\_Directional]}) He fell down on the ground. (IIC \tb{[Spatial\_Contact]})   He landed on his arm and broke it. (IID \tb{[Kinship]}) His mother brought him to the emergency room.  (IIE \tb{[Causation]})   It caused him to fall down the stairs. \\
                        \vspace{1pt}
                          \textbf{U:} \taub{Infer candidates before} (II) She hired a lawyer.  \\
                          \textbf{S:} (IA \tb{[Submitting\_Documents]}) She filed a civil lawsuit. (IB \tb{[Trial]}) She sued for breach of contract. (IC \tb{[Personal\_Relationship]}) She filed for divorce.  (ID \tb{[Awareness]}) She didn't know how to defend herself. (IE \tb{[Desiring]}) she did not want to go to jail. 
                  \\ \midrule  
                    \textbf{D. Counterfactual Story Rewriting} \\ \cmidrule(lr){1-1}
                    \textbf{User:} (I) Alec's daughter wanted more blocks to play with. (II) Alec figured that blocks would develop her scientific mind. (III) Alec bought blocks with letters on them. (IV) Alec’s daughter made words with them rather than
                    structures. (V) Alec was happy to see her developing her verbal ability. \\
                    \taub{Replace (II) with ``Alec could not afford to buy new blocks for his daughter'' and rewrite the last three sentences.} \\
                    \textbf{Parser:} (III) \tb{[Containers]} (IV) \tb{[Text\_Creation]} (V) \tb{[Emotion\_directed]} \\
                    \textbf{System:} {(I) Alec's daughter wanted more blocks to play with.} (II) Alec could not afford to buy new blocks for his daughter. \textit{(III) Alec's daughter begged him to buy her blocks.} \textit{(IV) Alec wrote a letter to Santa Claus himself.} \textit{(V) She was very happy when he wrote back.}
            \end{tabular} 
        } \\
    \bottomrule
\end{tabular}
\caption{Example use cases of frame-guided infilling. \textbf{A.} depicts \textit{human-in-the-loop iterative story refinement}, in which a user provides an initial context and/or intended frame semantic content and interacts with the model to predict and user-select new frame content and surface-realized context. \textbf{B.} depicts surface realization from a \textit{frame semantic story skeleton}, i.e. a seed sentence and a sequence of frame sets to appear in the specified order. \textbf{C.} depicts \textit{semantically diverse candidate generation}
using model frame inference to identify distinct semantic content then using conditional generation to realize each candidate. \textbf{D.} depicts \textit{counterfactual story revision},
in which one sentence (II) is replaced and subsequent sentences are rewritten using frames parsed from the originals. }
\label{fig:interactive}
\end{figure*}
From these results we can conclude that in the process of achieving effective controlled frame-guided language generation, the fine-tuned FFL model achieves competitive performance to its unconstrained ILM counterpart, especially in the presence of increased guiding information. Moreover, the compromise in quality for the LCD method is minimal particularly for single frame guidance.

\paragraph{Effect of Different Levels of Guidance}
\autoref{tab:indistinguish} and \autoref{tab:rel-plau-rank} show that as the level of guidance (number of frames provided) increases, FFL and LCD models show opposite trends in quality: the former improves whereas the latter gets worse. We illustrate this effect in \autoref{fig:ffl_lcd_trend}.

For FFL, this indicates that generative capabilities would improve if the model were trained with more information about semantic content. This is a somewhat counterintuitive finding, given the effectiveness of the ILM model trained with no semantic information whatsoever beyond surface-level lexical information (words in the context).  

For LCD, we posit that the increase in the size of lexical unit constraint sets amplifies the negative effects of the lexical units' word sense ambiguity, resulting in the downward trend. With more guiding frames, LCD has to search through a larger space of possible LU combinations and is therefore more prone to the misuse of LU (sense). 
Moreover, we observe that in some cases with many (e.g. $\geq 5$) frames, LCD cannot satisfy all constraints within one sentence
and will start new sentences to complete unmet constraints. This is likely a contributing factor to LCD's lower scores 
under human evaluations. 

\section{Case Study: Interactive Generation}
To demonstrate the practical applicability of our frame-guided infilling methods, we qualitatively explore them in a variety of human-in-the-loop use cases based on recent work in text generation. 
In the following cases, we use models for both frame ID inference and text infilling conditioned on surrounding context. For frame inference, we use the forward frame token probability of an unordered-frame M-FFL model trained as in Section 3, with the modification that training examples have between 0 and 4 surrounding sentences as context. This allows for more flexibilty than a model trained only on complete 5-sentence stories. 
We modify the training data by taking a random contiguous slice of each 5-sentence example. 
\autoref{fig:interactive} shows examples of each scenario.
For infilling, we use FFL for \textbf{A}, \textbf{B} and \textbf{D} and LCD for \textbf{C}. 

\paragraph{A. Iterative Story Refinement}
For a maximally free-form and extensible use case, we devise a scenario in the spirit of \newcite{goldfarb-tarrant-etal-2019-plan} in which a user interfaces with a model to collaboratively construct an open-domain story given any combination of text and/or frames. 
Over the course of a human-system dialog, the user can iteratively either choose for the model to predict new frames at specified locations in the context or select from candidate infills conditioned on selected frames. As discussed in \newcite{goldfarb-tarrant-etal-2019-plan}, this type of process allows for a symbiotic relationship in which the user can correct, suggest or revise content generated by the machine and vice versa. Injecting frame guidance into this scenario enables for an extra degree of interactive flexibility in both suggestion and specification. 

\paragraph{B. Generation from Story Skeleton}
Recent work~\cite{fan2018hierarchical,goldfarb-tarrant-etal-2019-plan} has used pretrained neural language models for surface realization of structured story content. We approximate this task by having a model accept a seed sentence (i.e. a prompt) plus an ordered sequence of sets of frames specifying the content to appear in a story. We then use the frame-guided conditional generation to complete the text. Without the ability to handle explicit frame semantic guidance, this task would be incredibly difficult for a neural generation model. 

\paragraph{C. Diverse Candidate Generation}
\newcite{weir-etal-2020-cod3s} explore the task of diverse causal generation, in which a model must propose a set of semantically distinct causes or effects of an input sentence. Following their two-step approach, we devise a frame semantic model that 1) predicts the distinct frames that are likely to appear at a specified index before (for causes) or after (effects) the input sentence, then 2) run a separate beam search conditioned on each top-$k$ predicted frame. Using a frame-infused generation model for this purpose leverages the hierarchical semantic delineations contained within FrameNet, selecting human-interpretable semantic spaces from which to generate content. This is compared to other methods for diverse sampling, such as random and nucleus sampling~\cite{holtzman2019curious}, in which there is no notion of higher level semantic reasoning and a tendency to hallucinate content, or COD3S~\cite{weir-etal-2020-cod3s}, which enables only moderate interpretability not based--as FrameNet is--in cognitive theories of semantic organization.

\paragraph{D. Counterfactual Story Revision}
\newcite{qin-etal-2019-counterfactual} introduce the task of generative counterfactual reasoning in narratives. Given an original story and a counterfactual event (i.e. the replacement of one original sentence), the task is to minimally revise the rest of the story according to the counterfactual replacement. We devise a frame semantic model for this task that 1) parses the frames of sentences following the replacement and 2) conditions the generation model on the replacement text and a sampled sequence of the parsed frames so as to produce a revised story whose frame semantics are similar to the original's. While previous approaches to this generation task condition only on surrounding context, our frame-injected model allows for explicit retention of semantic spaces.
\section{Conclusion}
We propose the application of frame semantics in the context of controlled text generation.
We introduce two extensions of neural text generation that leverage FrameNet frames as guiding signals: 1) model fine-tuning with a frame-guided infilling objective; and 2) disjunctive lexically constrained decoding with frame-associated lexical units. Experimental results on a sentence infilling task and the case study involving an interactive story generation setup show that both of our methods can properly leverage the frame information to trigger surface realization of frame semantic content. 
Our results show that our methods enable explicit manipulation of semantics at the frame level with competitive generation quality, and we exhibit a variety of use cases that enable new dimensions of user guidance on generation.

\section*{Acknowledgments}
This work was supported by DARPA KAIROS and NSF grant no. BCS-2020969. The U.S. Government is authorized to reproduce and distribute reprints for governmental purposes. The views and conclusions contained in this publication are those of the authors and should not be interpreted as representing official policies or endorsement.

\bibliography{anthology,framefilling}
\bibliographystyle{acl_natbib}

\appendix
\clearpage
\section{Training Details}
\subsection{FFL}
\label{app:training-appendix}
We finetune GPT-2 on examples of frame-guided infilling using the same training parameters (to the extend possible) as \citet{donahue2020enabling}. We use the \texttt{fairseq} library to perform training and inference using the pretrained GPT-2 parameters provided by HuggingFace\footnote{\url{https://github.com/pytorch/fairseq/blob/master/fairseq/models/huggingface/hf_gpt2.py}}. Training takes 1.5 hours using 8 Quadro RTX 6000 GPUs. \\
\begin{verbatim}
fairseq-train 
--task framefilling 
--sample-break-mode eos 
--arch ilm_gpt2 
--dropout 0.1 
--attention-dropout 0.1 
--clip-norm 1 
-optimizer adam --adam-eps 1e-08 
--lr 5e-5
--weight-decay 0.0
--max-epoch 100 
--patience 3 
\end{verbatim}
\subsection{ILM}
\label{app:ILM}
To compare ILM with FFL on a uniform basis, we retrain ILM on sentence level infilling using the code provided by \newcite{donahue2020enabling},\footnote{\url{https://github.com/chrisdonahue/ilm}} with same parameters and stopping criterion. 

It is worth noticing that the original ILM is trained on stories from the ROCStories dataset with titles provided. However, the test set portion of ROCStories on which we formulate the frame-guided sentence infilling task are provided without title. We observe that the original ILM trained with title is problematic in infilling the first sentence of a story without title (Sometimes it outputs full stop only, or generate a new title in addition to the sentence). Therefore, we delete all titles in the training data when retraining ILM.

\section{LCD Diversification}
Although the LCD algorithm will explore the prefix of each of the dozens of constraints typically associated with a frame, a few LUs will tend to dominate the final candidates throughout beam search --- this is also observed in \newcite{li2020guided}. 
This problem is exacerbated by the rather broad definitions of some frames that cover both general, common LUs, and more specific LUs, whose likelihood will be dwarfed during decoding by the former.
For example, the \textbf{Collaboration} frame contains LUs that depict the concept of collaboration from various perspectives: the act of collaborating (e.g.\ \textit{collaborate.v}, \textit{team up.v}), the participants in the collaboration (e.g.\ \textit{collaborator.n}, \textit{partner.n}), and the state of being in collaboration (e.g.\ \textit{in cahoots.a}, \textit{together.adv}), etc. However, in practice the general unit \textit{together.adv} is more often selected by beam search to satisfy the constraint because of its generally higher likelihood. This dominant LU prevents other potentially diverse surface realizations of the frame triggered by other LUs.

To improve the lexical and semantic diversity in triggering frames, we construct disjunctive sets on a more fine-grained semantic level. We divide each set of LUs into $k$ subsets using hierarchical clustering over the GloVe embeddings of LUs \cite{pennington2014glove}. In particular, we use the \texttt{AgglomerativeClustering} class of \textbf{scikit-learn} \footnote{\url{https://scikit-learn.org/stable/modules/generated/sklearn.cluster.AgglomerativeClustering.html}} to perform hierarchical clustering over the GloVe embedding of LUs to divide each set of frame-associate LUs into subsets. In the experiments, we set number of clusters to 8. For multi-frame constraints, we set number of clusters to 4 for the frame with the most number of LUs and 2 for the frame with the second most of, we do not divide any LU sets for remaining frames (if any), this could ensure the total combination of multi-frame LU subsets equals  8.
\autoref{fig:clusters} shows the clustering results of three frames: \textbf{Collaboration}, \textbf{Ingestion} and \textbf{Departing}, with number of clusters set to 4.

To ensure that the decoder will be able to explore all possible combinations of LUs, we build lists of tries for every combination of LU subsets. The constrained beam search is then run separately on each of them. 
To ensure that candidates from each LU subset are considered, final candidates are selected in a round-robin manner: the top-1 scored hypothesis is picked for each subset, followed by the top-2, and so on.

\begin{figure}[t]
\tiny
    \centering
    \begin{tabular}{l|l}
    \toprule
    \textbf{Frames} & \textbf{LUs clusters} \\
    \textbf{Collaboration} & \textbf{cluster 1}: conspire, conspiracy, collusion, collude \\
    \multirow{ 2}{*}{} & \textbf{cluster 2}: together, in league, in cahoots,\\ 
    & work together, team up \\
    & \textbf{cluster 3}: confederate \\
    \multirow{ 2}{*}{} & \textbf{cluster 4}: partner, jointly, cooperation, associate, affiliated, \\ 
    & collaboration, collaborator, cooperate, collaborate \\
    \hline
    \textbf{Ingestion} & \textbf{cluster 1}: have, put away, lap, put back, down \\
    & \textbf{cluster 2}: feed, lunch, breakfast, snack, eat, drink\\
    \multirow{ 3}{*}{} & \textbf{cluster 3}: swig, ingestion, quaff, swill, guzzle, \\
    & sup, nosh, gulp, devour, gobble, ingest, \\ 
    & consume, dine, nibble, imbibe, slurp, sip \\
    & \textbf{cluster 4}: tuck, munch, feast, nurse\\ 
    \hline
    \textbf{Departing} & \textbf{cluster 1}: departure, depart, exit, leave\\
    & \textbf{cluster 2}: vamoose, decamp, skedaddle \\
    & \textbf{cluster 3}: exodus, disappearance, escape \\
    & \textbf{cluster 4}: disappear, vanish, emerge \\
     \bottomrule
    \end{tabular}
    \caption{clustering examples of frame \textbf{Collaboration}, \textbf{Ingestion}, and \textbf{Departing}, morphological variants are excluded for demonstration purpose.}
    \label{fig:clusters}
\end{figure}

\section{Perplexity}
\label{app:ppl}
We repeat the perplexity experiment from \autoref{sec:auto-eval}, but instead of masking one out of five of a story's sentences at a time, we mask all five. This scenario can be considered a fully generative model of text in which no context is provided except for frame IDs specifying general semantic content for each sentence. \autoref{tab:ppl-all-masked} shows the resulting model perplexities.
\section{Human Evaluation Details}
\label{app:human-eval}
Akin to \citet{donahue2020enabling}, we sampled 100 stories from the test set of the ROCStories dataset. Masking one sentence at a time in each 5-sentence story, we obtained 500 masked stories. Each model was then tasked to infill a missing sentence in a masked story. We compared 10 models in total: 8 proposed in this paper (S/M/A-FFL, M/A-FFL-ordered, and the ordered variant\footnote{Based on the Frame Fidelity and the pilot HIT results, we chose to only evaluate the ordered variant, as the unordered LCD performed very similarly in terms of those metrics.} of S/M/A ILM+LCD), as well as the gold human infill and the ILM model. Below we further specify the details of each of the human evaluation tasks.

\subsection{Indistinguishability}
\begin{figure*}[t]
\centering
\includegraphics[width=0.90\textwidth]{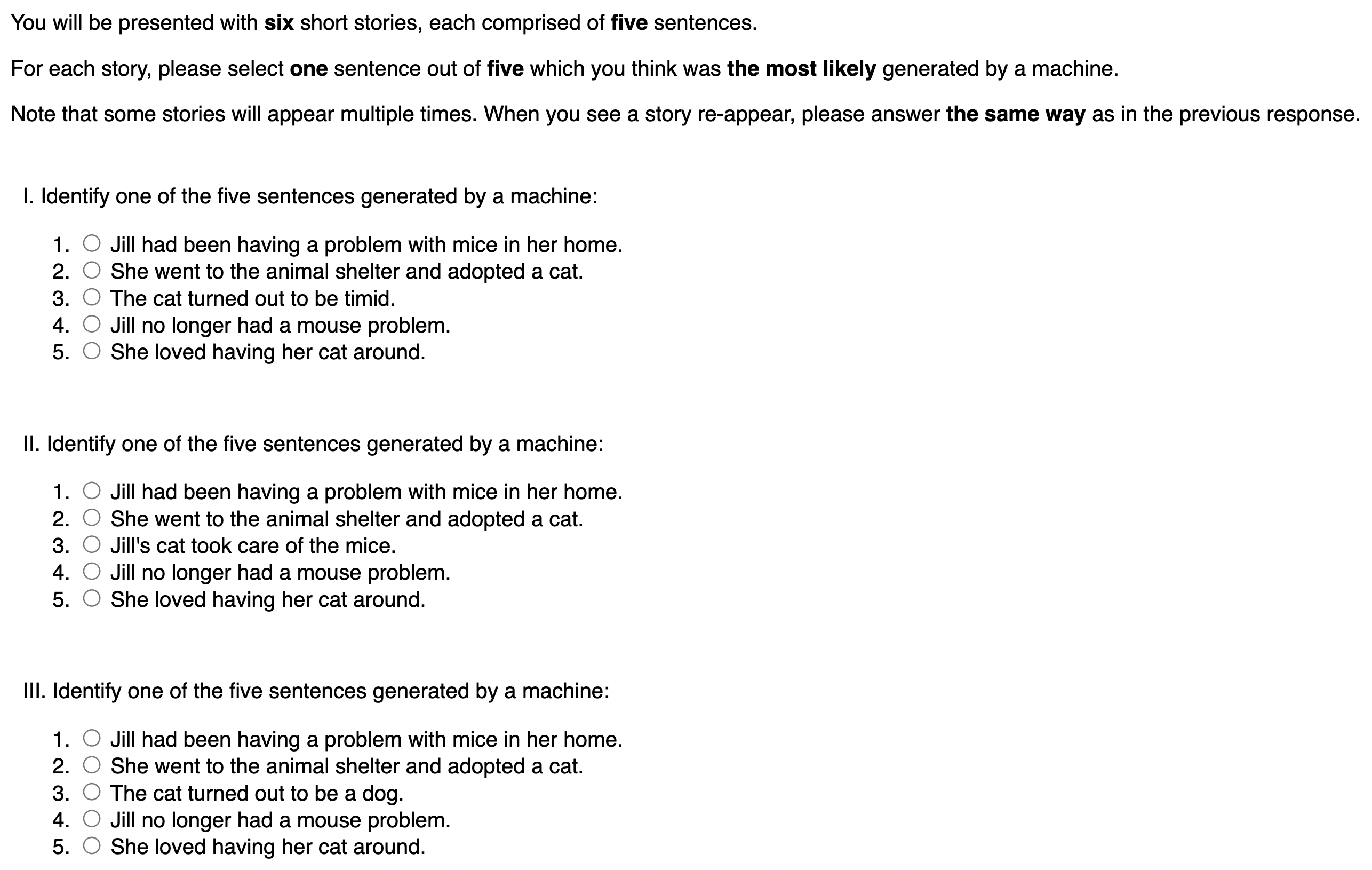}
\caption{Interface shown to the workers during collection of indistinguishability judgments.}
\label{fig:turing-test-ui}
\end{figure*}

To achieve high comparability with \citet{donahue2020enabling}, we conducted this evaluation as a Human Intelligence Task (HIT) on Amazon Mechanical Turk. To filter out malicious workers, we used a control model which always generates ``This sentence was generated by a machine.'' or a synonymous sentence. We also validated that the gold human infill achieves 80\% confusion rate (which was attained precisely in our run), which corresponds to picking 1 sentence out of 5 at random.  Overall, 12 workers participated in the HIT, of which one was filtered by the control model. The annotator's interface can be seen on \autoref{fig:turing-test-ui}.

\begin{table}[t!]
    \small
    \centering
    \setlength\tabcolsep{3pt}
    \begin{tabular}{lrrrrr} \toprule
    \multirow{2}*{\textbf{Perplexity}} & \multirow{2}*{ILM} & \multirow{2}*{S-FFL} & \multirow{2}*{M-FFL} & \multirow{2}*{A-FFL} & A-FFL  \\ &&&&& (ord) \\ \midrule
        Infill Text  & 13.88  &  11.07 & 8.76 & 5.45 & \textbf{4.69} \\
        \quad + Sp Toks & 8.87 & 9.16 & 10.05 & 9.3 & \textbf{7.43}             \\
        \quad + 5 Fr Slots & \textbf{4.66} & 5.51 & 6.71 & 7.64 & 6.23 \\
        \bottomrule
    \end{tabular}
    \caption{Model perplexity over infill text tokens and infill text tokens + special tokens with all 5 ROCStory sentences masked out.}
    \label{tab:ppl-all-masked}
\end{table}

\subsection{Relative Plausbility}
\begin{figure*}[t]
\centering
\includegraphics[width=0.90\textwidth]{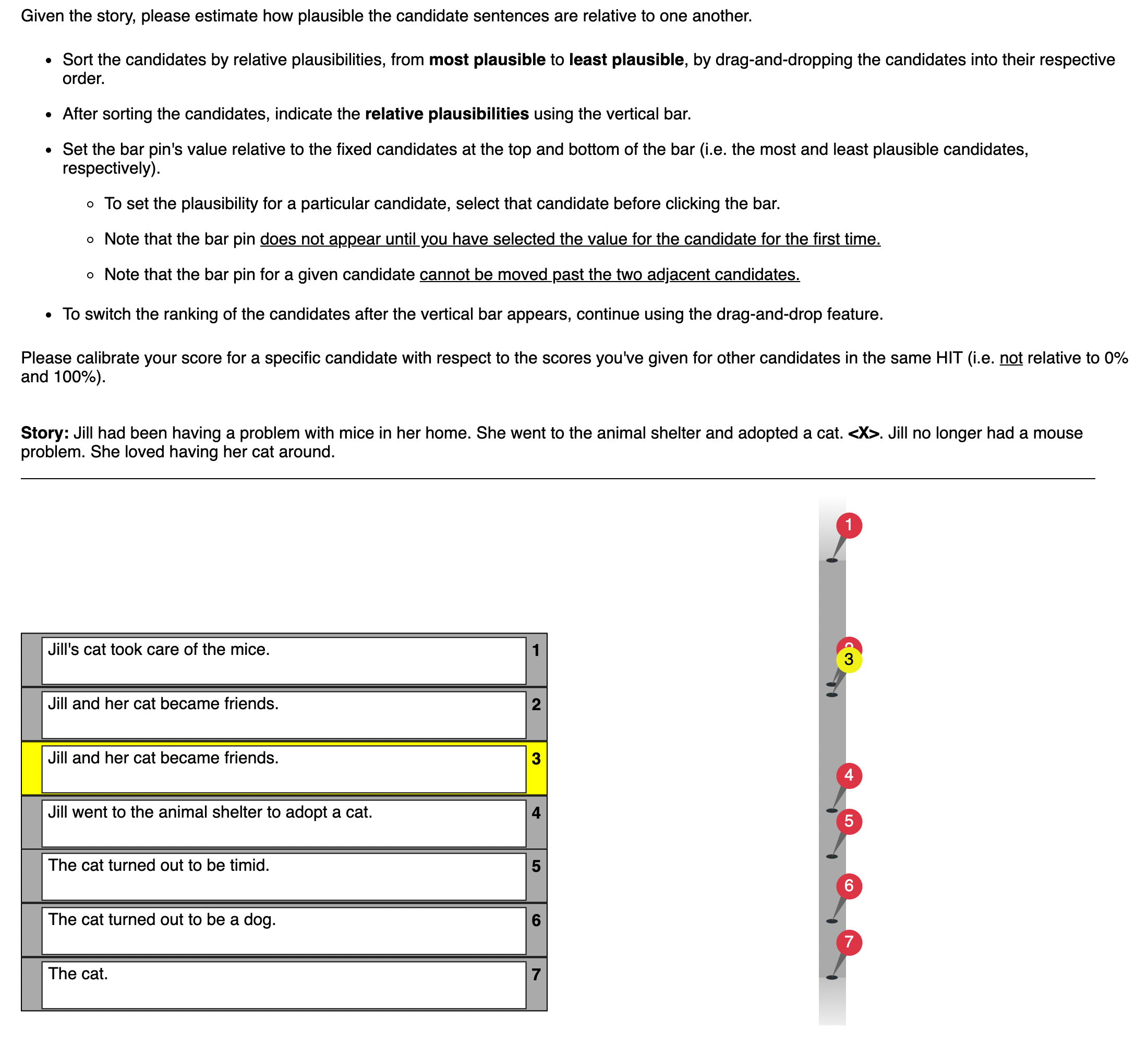}
\caption{Interface shown to the workers during collection of relative plausibility judgments.}
\label{fig:rel-plaus-ui}
\end{figure*}

Due to a relatively high complexity of this task, compared to the Indistinguishability task, the evaluation was conducted with a team of skilled annotators, comprised of four undergraduate students who have previously participated in NLP/AI annotation projects.
On average, ranking 10 models' outputs for one story took 3 minutes 19 seconds for each worker. The annotator's interface can be seen on \autoref{fig:rel-plaus-ui}.


\end{document}